\documentclass[10pt,twocolumn,letterpaper]{article}

\usepackage{wacv}              % To produce the CAMERA-READY version
\usepackage{graphicx}
\usepackage{amsmath}
\usepackage{amssymb}
\usepackage{booktabs}

\usepackage{multirow}
\usepackage{enumitem}
\usepackage{bbm}
\usepackage{wrapfig}
\usepackage{setspace}
\usepackage{footnote}
\usepackage{xcolor}         % colors
\usepackage{amssymb}
\usepackage{tabularx}
\usepackage{colortbl} % for colored tables
\usepackage{pifont}  % for check and cross marks
\usepackage{cite}  % to keep cites numerical order
\usepackage{mathtools}  % for ceiling and flooring
\usepackage[font=small]{caption}  % caption style
\usepackage{blindtext}  % for dummy text generation
\usepackage{stmaryrd} % short arrows
\usepackage[accsupp]{axessibility}  % Improves PDF readability for those with disabilities.
\usepackage{mathtools}
\usepackage{subfiles}
\usepackage{afterpage}

\usepackage{algorithm}
\usepackage{algpseudocode}
\usepackage{adjustbox}
\usepackage{anyfontsize}

\makeatletter
\newcommand{\multiline}[1]{%
  \begin{tabularx}{\dimexpr\linewidth-\ALG@thistlm}[t]{@{}X@{}}
    #1
  \end{tabularx}
}

\newcommand\blfootnote[1]{%
  \begingroup
  \renewcommand\thefootnote{}\footnote{#1}%
  \addtocounter{footnote}{-1}%
  \endgroup
}

\definecolor{amethyst}{rgb}{0.6, 0.4, 0.8}

\definecolor{grey}{rgb}{0.9, 0.9, 0.9}
\definecolor{grey_green}{RGB}{237 249 230}

\newcommand{\ccol}{\cellcolor{grey_green}}
\newcommand*\colourcheck[1]{%
  \expandafter\newcommand\csname #1check\endcsname{\textcolor{#1}{\ding{51}}}%
}

\colourcheck{blue}
\colourcheck{green}
\colourcheck{red}
\newcommand{\xmark}{\ding{55}}%

\def\ie{\emph{i.e.}}

\definecolor{bright_red}{rgb}{0.97, 0.9, 0.9}
\definecolor{blk}{rgb}{0, 0, 0}
\definecolor{grn}{rgb}{0, 0.6, 0}
\definecolor{mgt}{rgb}{0.8, 0.1, 0.8}
\definecolor{darkblue}{rgb}{0.2, 0.2, 0.8}
\definecolor{orange}{rgb}{1.0, 0.5, 0.2}
\definecolor{goldenrod}{rgb}{0.85, 0.65, 0.13}

\newcommand{\sytodo}[1]{{\color{grn}{#1}}}

\newcommand{\newred}[1]{{\color{red}{#1}}}

\usepackage[pagebackref,breaklinks,colorlinks]{hyperref}

\usepackage[capitalize]{cleveref}
\crefname{section}{Sec.}{Secs.}
\Crefname{section}{Section}{Sections}
\Crefname{table}{Table}{Tables}
\crefname{table}{Tab.}{Tabs.}

\def\wacvPaperID{01919} % *** Enter the WACV Paper ID here
\def\confName{WACV}
\def\confYear{2025}

\begin{document}

\title{Learning Unified Distance Metric Across Diverse Data Distributions \\ with Parameter-Efficient Transfer Learning}

% \and

\author{Sungyeon Kim$^1$ \hspace{10mm}  
Donghyun Kim$^{*2}$ \hspace{10mm} 
Suha Kwak$^{1}$ \\
$^1$Pohang University of Science and Technology~(POSTECH) \hspace{5mm} $^{2}$Korea University \\
{\tt\small \{sungyeon.kim,suha.kwak\}@postech.ac.kr, d{\_}kim@korea.ac.kr} \\
}

\maketitle
\blfootnote{$^*$ The work was done when Donghyun Kim worked at MIT-IBM Watson AI Lab.}

%%% ABSTRACT %%%%%%%%%%%%%%%%%%%%%%%%%%%%%%%%%%%%%%%%%%%%
\begin{abstract}
A common practice in metric learning is to train and test an embedding model for each dataset. This dataset-specific approach fails to simulate real-world scenarios that involve multiple heterogeneous distributions of data. In this regard, we explore a new metric learning paradigm, called Unified Metric Learning (UML), which learns a unified distance metric capable of capturing relations across multiple data distributions. UML presents new challenges, such as imbalanced data distribution and bias towards dominant distributions. These issues cause standard metric learning methods to fail in learning a unified metric. To address these challenges, we propose Parameter-efficient Unified Metric leArning (PUMA), which consists of a pre-trained frozen model and two additional modules, stochastic adapter and prompt pool. These modules enable to capture dataset-specific knowledge while avoiding bias towards dominant distributions. Additionally, we compile a new unified metric learning benchmark with a total of 8 different datasets. PUMA outperforms the state-of-the-art dataset-specific models while using about 69 times fewer trainable parameters.
\end{abstract}

%%% INTRODUCTION %%%%%%%%%%%%%%%%%%%%%%%%%%%%%%%%%%%%%%%%
% \input{_1_introduction.tex}
% \vspace{-2mm}
% \begin{figure} [!h]
% % \vspace{-5mm}
% \centering
% \includegraphics[width = 1.0\linewidth]{figs/Figure_thumbnail_new.pdf}
% \vspace{-5mm}
% \caption{
% We introduce PUMA, a universal metric learning approach that aims to learn a unified distance metric capable of capturing semantic similarity across multiple data distributions. Unlike the existing metric learning approach, PUMA facilitates image retrieval across diverse heterogeneous distributions. It consistently achieves high performance compared to existing models devoted to each dataset and CLIP, trained on extensive data.
% } 
% \label{fig:thumbnail}
% % \vspace{-3mm}
% \end{figure}

\section{Introduction}
\label{sec:intro}

% Learning semantic distance metrics has been playing a key role in machine learning applications including visual recognition~\cite{bell2020groknet, tang2019msuru}, retrieval system~\cite{yang2017visual, huang2020embedding, zhang2018visual, zhai2019learning}, recommender systems~\cite{hou2022towards, ma2020probabilistic}, and representation learning~\cite{kim2019deep, he2020momentum, chen2020simple}. 
% Deep metric learning has served as a representative approach to learning semantic distance metrics: it aims to learn highly nonlinear distance metrics through deep neural networks that approximate the actual underlying semantic similarity between samples.

\begin{figure*} [!t]
% \vspace{-2mm}
\vspace{-1mm}
\centering
\includegraphics[width = 0.94\linewidth]
{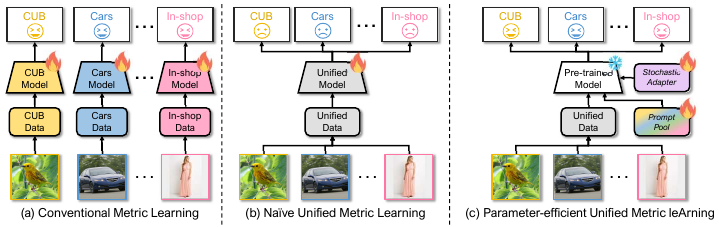}
\vspace{-2mm}
\caption{\textbf{Comparison between conventional and unified metric learning methods.} (a) Conventional metric learning employs separate models for individual datasets, incurring significant computational and memory costs as data diversity grows. (b) A na\"ive solution is to fine-tune the model on a merged dataset, but this often leads to a severe bias toward major data distributions. (c) In contrast, our method excels on all datasets with just one model. This is highly resource-efficient as it enables one-time learning and evaluation on diverse data distributions using a single model.}
\label{fig:comparison}
\vspace{-3mm}
\end{figure*}

Deep metric learning has emerged as a representative approach in machine learning, particularly effective in understanding semantic distances between data samples. By harnessing the capabilities of deep neural networks, this approach enables learning highly nonlinear distance metrics that approximate the actual underlying semantic similarity between samples. It plays a key role in a wide array of applications, including visual recognition~\cite{bell2020groknet, tang2019msuru}, retrieval system~\cite{huang2020embedding, zhang2018visual}, recommendation~\cite{hou2022towards, ma2020probabilistic}, and representation learning~\cite{kim2019deep, he2020momentum, chen2020simple}.
% Notably, this approach underpins the technology behind visual search platforms such as Bing Visual Search~\cite{hu2018web}, Pinterest's Flashlight and Lens~\cite{zhai2019learning}, and eBay Visual Shopping~\cite{yang2017visual}, illustrating its widespread influence across various industries.
While metric learning methods have achieved remarkable progress, they typically focus on learning metrics within a single domain. However, real-world applications often violate this assumption and involve multiple heterogeneous data distributions. For instance, users of a retrieval system may query data with substantially different semantics and diverse distributions. 
{A na\"ive solution to this issue is training multiple models for each of the different data distributions and combining them through ensemble techniques or toggling between them based on the query.}
% To address this issue, conventional approaches might train multiple models to handle the different data distributions, combining them through ensemble techniques or toggling between models based on the query. 
{This process, illustrated in Fig.~\ref{fig:comparison}(a), demands extensive computational resources to find the most effective model or combination for each data distribution.}
% This process, illustrated in Fig.~\ref{fig:comparison}(a), requires extensive computational resources and involves a grid search of various methods and parameters to find the most effective approach for each data distribution.

Recent efforts in metric learning have aimed at developing unified embedding solutions capable of generalizing to multiple fine-grained object types with a single model~\cite{uned, almazan2022granularity}, ensuring scalability in real-world scenarios. This approach, which we refer to as \textbf{Unified Metric Learning~(UML)}, seeks to learn a unified distance metric that captures semantic similarity across multiple data distributions. UML involves training a single model on a composite dataset derived from various sources, without labels indicating their origin. This mirrors real-world scenarios where data are collected from diverse sources and a model needs to generalize across heterogeneous data distributions.

While previous work~\cite{uned, almazan2022granularity}, has made strides in this direction by providing datasets, it has not fully addressed or analyzed the challenges posed by UML. Our study reveals that these challenges extend beyond simply aggregating diverse data sources. One major challenge is the \emph{imbalance in the quantity of data from different sources}. This imbalance arises naturally due to the difference between data sources in terms of accessibility and annotation cost. Our analysis shows that na\"ive fine-tuning on combined data from multiple sources similar to the approach used in previous work\cite{uned}, as depicted in Fig.~\ref{fig:comparison}(b), results in models that are strongly biased toward the sources with larger amounts of data.
Another challenge is \emph{the variability of key discriminative features across data sources}. For instance, color may be crucial for distinguishing bird species but irrelevant or even misleading for differentiating vehicle types. Therefore, models should learn to capture specific discriminative features from each data source to achieve effective UML.

To address these challenges, we propose a novel approach called \textbf{Parameter-efficient Unified Metric leArning~(PUMA)}, which is a completely different direction from existing metric learning. While most existing metric learning approaches focus on developing new metric designs, PUMA addresses UML from an architectural perspective. PUMA aims to capture unified semantic similarity through a single embedding model while mitigating the imbalance in data quantity across sources.
To achieve this, we draw inspiration from recent advances in parameter-efficient fine-tuning ~\cite{houlsby2019parameter, pfeiffer2020adapterfusion, he2021towards, li2021prefix}.
% from Natural Language Processing (NLP) domain. 
%
% Our key idea is to freeze the parameters of a model pre-trained on a large-scale dataset, thereby preserving its generalization capability, and learning dataset-specific knowledge from the unified dataset with a minimal number of additional parameters.
Our key idea is to freeze the parameters of a model pre-trained on a large-scale dataset, thereby preserving its generalization capability. We then introduce two modules: one that learns common knowledge across all sources without 
%biasing towards any specific one, 
{being biased towards any specific data source}
% and another 
{and the other} 
that learns source-specific knowledge with minimal extra parameters.

Specifically, PUMA is built on a pre-trained Vision Transformer (ViT) and incorporates two additional modules, namely \emph{stochastic adapters} and a \emph{prompt pool} (see Fig.~\ref{fig:comparison}(c)).
Stochastic adapters are lightweight modules that learn knowledge spanning across all data sources and operate in parallel with the corresponding transformer block. Their operation is randomly switched off during training, preventing bias towards specific sources by randomly providing adapted features or pre-trained model features. This approach is parameter-efficient and improves performance across diverse data distributions.
On the other hand, the prompt pool builds conditional prompts accounting for distinct characteristics of each data distribution on the fly. Organized in a key-value memory structure, the prompt pool dynamically generates conditional prompts based on input by aggregating relevant prompts using an attention mechanism. These prompts are added to the input sequence, allowing for representing the unique properties of data source.

% Meanwhile, the prompt pool is used to build a conditional prompt that accounts for distinct characteristics of each data distribution on the fly. 
% Specifically, the prompt pool is a set of prompts organized in a key-value memory, and given the input feature, the conditional prompt is generated by aggregating relevant prompts in the pool using an attention mechanism.
% The conditional prompt is added to the input sequence of the ViT, allowing for more targeted adaptation.
To comprehensively evaluate UML approaches, we compile a benchmark comprising eight datasets from various domains and classes. This benchmark incorporates standard metric learning datasets~\cite{CUB200,krause20133d, songCVPR16, DeepFashion}, enabling direct comparisons with existing metric learning methods~\cite{
Schroff2015, sampling_matters, kim2020proxy, deng2019arcface, ren2024learning}
while also extending the evaluation to new domains. 
% We compile a UML benchmark comprising eight datasets from different domains and classes. 
Our single model significantly outperforms those trained on combined data from multiple sources using conventional metric learning techniques.
Remarkably, it surpasses most models trained individually on each dataset, achieving this with 69 times fewer trainable parameters. 
% Additionally, we demonstrate that our method serves as a strong few-shot learner.
The contribution of this paper is three-fold:
\vspace{-2mm}
\begin{itemize}[leftmargin=5mm]
    \setlength\itemsep{0.2ex}
    % \item We first present UML, a new paradigm for learning a unified distance metric that captures semantic similarity across diverse data distributions.

    \item We propose PUMA, a novel method utilizing stochastic adapters and a prompt pool to address data quantity imbalance and feature variability in UML, respectively.

    \item We establish a new benchmark for UML with eight diverse datasets and investigate conventional metric learning methods with loss variants and existing parameter-efficient fine-tuning strategies on this benchmark.

    \item Extensive experiments demonstrate that PUMA outperforms state-of-the-art models, including those specialized for individual sources, using 69 times fewer trainable parameters. 
    % Moreover, our method proves to be highly effective in few-shot learning scenarios.
\end{itemize}

%%% RELATED WORK %%%%%%%%%%%%%%%%%%%%%%%%%%%%%%%%%%%%%%%%
% \vspace{-2mm}
\section{Related Work}

\noindent\textbf{Deep Metric Learning.} It aims to learn a metric function that approximates the underlying semantic similarity of data by pulling semantically similar samples (positive) closer to the anchor and pushing dissimilar samples (negative) away. To achieve this goal, the development of loss functions has been the main focus of this field, typically classified into pair-based and proxy-based losses. Pair-based losses consider relations between pairs~\cite{sampling_matters, Bromley1994, Chopra2005, Hadsell2006}, triplets~\cite{Wang2014, Schroff2015} or higher-order tuples of samples~\cite{songCVPR16, Sohn_nips2016, wang2019ranked, wang2019multi, songCVPR17}. They can capture the fine-grained relations among samples, but they are suffering from the issue of increased training complexity as the number of training data increases. 
Proxy-based losses address the complexity issue by introducing learnable parameters called proxies to represent the training data of the same class. They greatly reduce the complexity of examining the relations between all data by considering those between data and proxies. In this direction, the approaches have used proxies to approximate the pair-based loss~\cite{movshovitz2017no, kim2020proxy, Qian_2019_ICCV} or have modified the cross-entropy loss~\cite{deng2019arcface, wang2018cosface, zhai2018classification, teh2020proxynca++}.
Existing metric learning methods have primarily focused on specific distributions within individual datasets. Recent work by~\cite{uned, almazan2022granularity} proposed datasets for unified metric learning, but their methods are limited to fine-tuned models via classification loss. Our work extends beyond this by addressing the unique challenges of UML through parameter-efficient techniques.

% adapter multi_domain,houlsby2019parameter,pfeiffer2020adapterfusion
% Lora hu2021lora
% fine-tuning subset sung2021training
% unified view of parameter - he2021towards
% prompt tuning - 
\noindent\textbf{Parameter-Efficient Fine-Tuning~(PEFT).}
Large-scale pre-trained models have shown significant improvements across various downstream tasks. As the model size and the number of tasks grow,  parameter-efficient fine-tuning approaches~\cite{hu2021lora,multi_domain,houlsby2019parameter,pfeiffer2020adapterfusion,he2021towards} have been developed to adapt to diverse downstream by updating only a small fraction/number of learnable parameters while fully utilizing the knowledge of the pre-trained model without catastrophic forgetting. Low-rank adaptation~\cite{hu2021lora} is proposed to approximate the parameter update or light-weight adapter modules~\cite{houlsby2019parameter,pfeiffer2020adapterfusion} can be inserted between pre-trained layers during fine-tuning. Prefix/prompt tuning~\cite{lester2021power,li2021prefix,wang2022learning,smith2022coda} has been introduced where additional learnable tokens (soft prompts) are added during fine-tuning while keeping the backbone frozen. 
In metric learning, there has been only one prior study~\cite{ren2024learning}, which tune prompts to generate semantic proxies. However, all existing PEFT methods are designed to handle only a single data distribution. We are the first to propose an efficient-tuning method that can deal with multiple data distributions in metric learning.
 
 % In contrast to prior deep metric learning work, we utilize parameter-efficient fine-tuning for our proposed universal metric learning setup to learn universal representations across different data distributions in a single model while preventing bias and catastrophic forgetting, which outperforms even full fine-tuning baselines.

%%% METHOD %%%%%%%%%%%%%%%%%%%%%%%%%%%%%%%%%%%%%%%%%%%%%%
\section{Unified Metric Learning}
   In this section, we first review conventional metric learning, and then introduce the UML setting and discuss its technical challenges.

\begin{table}[!t]
\setlength{\tabcolsep}{1pt}
\fontsize{6.6}{8}\selectfont
\resizebox{\linewidth}{!}{
\begin{tabularx}{1.0\linewidth}
    {
      p{0.18\linewidth}
      >{\centering\arraybackslash}X
      >{\centering\arraybackslash}X
      >{\centering\arraybackslash}X
      >{\centering\arraybackslash}X
      >{\centering\arraybackslash}X
      >{\centering\arraybackslash}X
      >{\centering\arraybackslash}X
      >{\centering\arraybackslash}X
      >{\centering\arraybackslash}X
      }
\toprule
& \multicolumn{1}{c}{\rotatebox[origin=c]{90}{CUB}} & \multicolumn{1}{c}{\rotatebox[origin=c]{90}{Cars}} & \multicolumn{1}{c}{\rotatebox[origin=c]{90}{SOP}} & \multicolumn{1}{c}{\rotatebox[origin=c]{90}{In-shop}} &  \multicolumn{1}{c}{\rotatebox[origin=c]{90}{NABirds}} & \multicolumn{1}{c}{\rotatebox[origin=c]{90}{Dogs}} & \multicolumn{1}{c}{\rotatebox[origin=c]{90}{Flower}} & \multicolumn{1}{c}{\rotatebox[origin=c]{90}{Aircraft}} & \multicolumn{1}{c}{\rotatebox[origin=c]{90}{Total}}\\ \midrule
{Train Samples} & 5.8K & 8.0K & 59.5K & 25.8K & 22.9K & 10.6K & 3.5K  & 5K & 141.4K  \\
{Train Classes} & 100 & 98 & 11.3K & 3.9K  & 278 & 60 & 51 & 50 & 15.9K  \\ \midrule
{Test Samples} & 5.9K & 8.1K & 60.5K  & 28.7K & 25.6K & 9.9K  & 4.7K  & 5K & 148.5K \\
{Test Classes} & 100 &  98 & 11.3K &  3.9K & 277 &  60 & 51 & 50 & 15.9K \\
\bottomrule
\end{tabularx}
}
\vspace{-2mm}
\caption{Dataset statistics: the number of training images and their classes used in training and testing.}
\label{tab:dataset_stats}
\vspace{-4mm}
\end{table}

\subsection{Revisiting Conventional Metric Learning}
\label{sec:problem_form}
Metric learning is the task of learning a distance function that captures the semantic dissimilarity between samples in a given dataset $S$. 
Such a distance function $d$ holds:
\begin{equation}
d(x, x^+; \theta) < d(x, x^-; \theta) \quad \forall (x, x^+, x^-), 
\label{eq:ML_constraint}
\end{equation}
where $x^+$ and $x^-$ denote the positive sample that belongs to the same class as $x$, and negative samples that are not, respectively, and $\theta$ represents the model parameters. Deep metric learning achieves this by learning a deep neural network as a high-dimensional embedding function,
% $f(\cdot,\theta)$, for individual data, 
and employing Euclidean or cosine distance to calculate the distance between embedding vectors. 

% \begin{figure} [!t]
% \includegraphics[width = 1.0\columnwidth]{figs/num_samples.pdf}
% \vspace{-7mm}
% \caption{The number of samples in each category. Each color represents its dataset. While the number of samples per class is small in SOP, it has a substantially greater number of samples compared to other datasets, which results in a bias towards large datasets.
% } 
% \label{fig:num_samples}
% \vspace{-3mm}
% \end{figure}

Note that metric learning seeks a generalization to classes unseen in training. The conventional setup thus employs a set of classes $C_{t}$ and their labeled samples $S_{t} = \{(x_t, y_t)~|~y_t \in C_{t}\}$ for training, and evaluates a trained embedding model for a set of unseen classes $S_{u} = \{(x_u, y_u)~|~y_u \in C_{u}\}$, where $C_{t}~\cap~C_{u} = \varnothing$ and $S_t~\cup~S_u = S$.
This convention only considers generalization within a single dataset.
% in which both training and test data are drawn from the same compact distribution.

\subsection{Problem Formulation of UML}
UML is an extension of conventional metric learning and tackles the challenging and practical problem of dealing with \textbf{multiple heterogeneous data distributions} using a single embedding model. The goal of UML is to learn a unified distance metric that can effectively capture diverse relations among samples across multiple distributions, 
while maintaining the intra-class compactness and inter-class separability within each distribution. 
In UML, a model is trained \emph{if it were dealing with a single dataset, unaware that the data originates from multiple sources}.
This mirrors real-world scenarios where data are collected from various sources, such as large-scale datasets from multi-modal distributions and a combination of multiple smaller datasets in practical applications.
% making it highly suitable for both a large-scale dataset drawn from a multimodal distribution and a union of multiple small datasets in real-world applications.

% In the UML setting, we assume 
To simulate this in practice, we use specific datasets to represent individual data distributions. 
% From this point on, 
For the brevity,
we refer to these individual data distributions or domains simply as \emph{datasets}. Suppose that we have $N_s$ datasets, denoted as $S^{1}, S^{2},\cdots, S^{N_s}$, and define the unified dataset $\mathbb{S} = \cup_{i=1}^{N_s} S^i$. To learn a unified embedding function, UML leverages the unified training dataset $\mathbb{S}_{t} = \cup_{i=1}^{N_s}{S^i_t}$, which aggregates training data from all the datasets. The learned unified distance function is evaluated in two different ways to assess its generalization capability. First, we evaluate it on the unified unseen test data $\mathbb{S}_{u} = \cup_{i=1}^{N_s}{S^i_u}$ to measure its universal accuracy, which demonstrates its capacity to comprehend semantic similarity across all datasets without favoring any specific dataset.
Second, we evaluate the distance function on the unseen test data of each dataset ${S^i_u}$ separately, to assess its ability to grasp the specific semantic similarity for each dataset.

\subsection{Challenges in UML}
UML encounters a new challenge -- \emph{dataset imbalance}, arising when integrating datasets of substantially different sizes, as shown in Table~\ref{tab:dataset_stats}. 
The imbalance occurs naturally due to the varying nature of data types and the difficulty of class labeling. For example, data of everyday objects are easier to collect and annotate compared to those of fine-grained object classes or those requiring expert knowledge. This leads to a disparity in the number of samples available in each dataset.
% \emph{highly imbalanced distribution of the unified dataset}. This dataset imbalance arises when integrating datasets of substantially different sizes, as shown in Table~\ref{tab:dataset_stats}. 
% This issue is particularly critical in metric learning because it results in a substantial portion of samples within each batch coming from the larger datasets. Consequently, the model tends to focus mainly on learning relations within these larger datasets, introducing a dataset bias.
In metric learning, this imbalance causes a substantial portion of samples in each batch to come from larger datasets, leading the model to excessively focus on relations on the dataset, thereby introducing a dataset bias. 
% Data imbalance is a common and well-known issue in a large variety of vision tasks. However, UML presents a more complex and unique challenge, where the entire data distribution becomes long-tailed due to class imbalance, and also has dataset imbalance caused by integrating datasets of substantially different sizes, as illustrated in Table~\ref{tab:dadtaset_stats}.
% This issue is particularly critical in metric learning:the dataset imbalance results in a substantial portion of samples within the batch from larger datasets, and consequently, the model tends to focus mainly on learning relations within these larger datasets, thereby introducing a dataset bias.

Another challenge in UML is that \emph{class-discriminative features are not shared across all datasets.} This challenge arises due to the disparity between different data distributions since each distribution has its own characteristics that define relations between its samples, which could conflict with those of the other distributions. For instance, while color may be crucial for differentiating between bird species, it may impede distinguishing between different vehicle types.
Thus, training with a unified dataset may lead to two potential problems. First, if the model focuses on class-discriminative features that are specific to a certain data distribution, it may harm datasets where those features are not relevant. Second, if the model attends to the commonalities shared among all datasets, its discriminability for capturing fine-grained differences between samples may diminish.

Moreover, UML still has a challenge in \emph{generalization to unseen classes}, inherited from conventional metric learning. However, this challenge becomes even more difficult as UML deals with diverse, imbalanced data distributions.
Adopting the traditional strategy of training multiple models and subsequently ensembling them is a straightforward way to handle such diverse datasets without confronting the above challenges.
However, this approach demands a vast number of parameters and substantial computational resources. Instead, we introduce a parameter-efficient approach that elegantly tackles all the challenges above.

\begin{figure*} [!t]
\centering
\includegraphics[width = 0.9\linewidth]{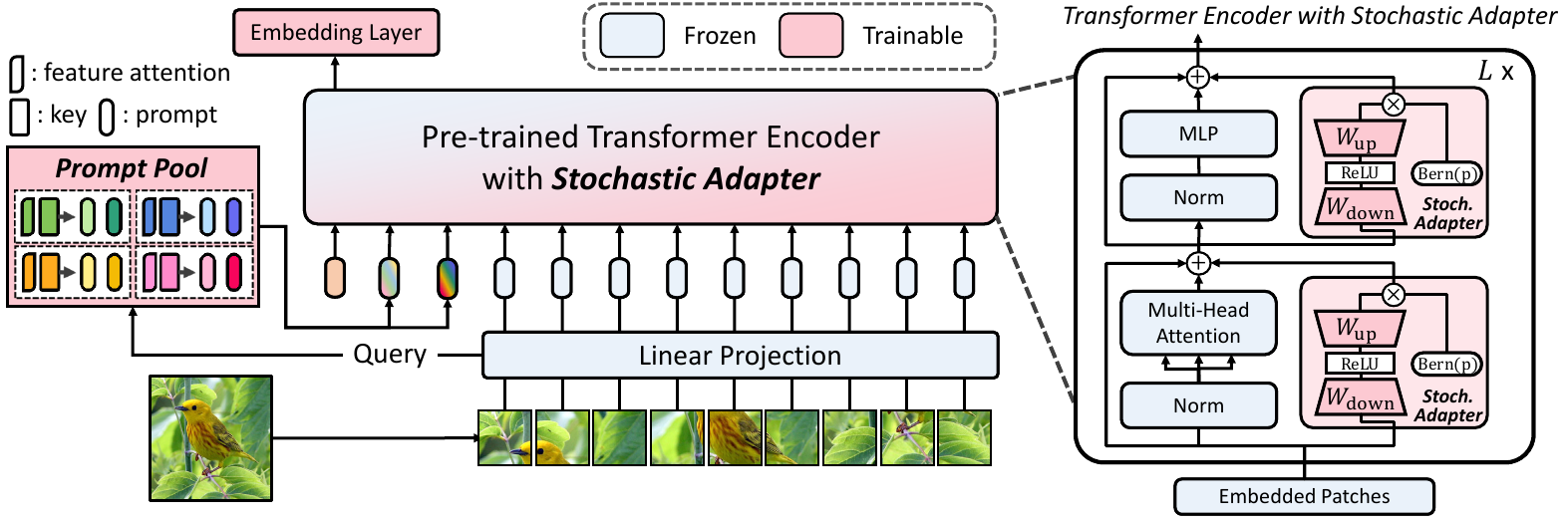}
\vspace{-2mm}
\caption{\textbf{An overview of PUMA.} PUMA consists of two learnable modules: stochastic adapters (Sec.~\ref{sec:adapter}) and a prompt pool (Sec.~\ref{sec:prompt}). Using the output of the transformer's embedding layer as a query, and it creates a conditional prompt by integrating relevant prompts through an attention mechanism. The conditional prompt is combined with image embeddings and class token, and then fed into the transformer. The modified input is embedded through transformer blocks, each coupled with a stochastic adapter, a learnable bottleneck module that turns on stochastically during training.} 
\label{fig:overview}
\vspace{-3mm}
\end{figure*}

\section{Proposed Method}
We propose a novel approach to UML, named Parameter-efficient Unified Metric leArning (PUMA). In contrast to conventional metric learning methods that fine-tune the entire model parameters, PUMA does not tune a large-scale pre-trained model but keeps its generalization capability across diverse data distributions. Instead, we leverage small additional modules that learn dataset-specific knowledge from the unified dataset. As shown in Fig.~\ref{fig:overview}, PUMA uses a pre-trained ViT as a backbone, and employs stochastic adapters and a prompt pool as the additional modules, which are detailed in the remainder of this section.

\subsection{Preliminaries: ViT}
\label{sec:vit}
ViT~\cite{ViT} is composed of a patch embedding layer and an encoder with $L$ sequential transformer layers. The patch embedding layer splits the input image $x$ into image patch embeddings $E \in \mathbb{R}^{N_e\times D}$, where $N_e$ denotes the number of patch embeddings and $D$ is the embedding dimension. The input sequence of the transformer encoder is formed by appending the image patch embeddings to a learnable class token embedding $e_\textrm{cls} \in \mathbb{R}^{D}$, as follows:
\begin{equation}
z_0 = [e_\textrm{cls}, E].
\label{eq:input_sequence}
\end{equation}
Each transformer layer consists of multi-headed self-attention (MSA) and multilayer perceptron (MLP) blocks, with layer normalization (LN) applied before every block and residual connections after every block:
\begin{equation}
\begin{aligned}
z'_{\ell} &= \textrm{MSA}(\textrm{LN}(z_{\ell-1})) + z_{\ell-1}, \quad &{\ell = 1, \dots, L}, \\
z_{\ell} &= \textrm{MLP}(\textrm{LN}(z'_{\ell})) + z'_{\ell}, \quad &{\ell = 1, \dots, L},
\label{eq:transformer_MSA}
\end{aligned}
\end{equation}

\subsection{Stochastic Adapter}
\label{sec:adapter}
To effectively adapt the model to the unified dataset without being biased to the large dataset. While adding learnable parameters shared by all data enables adaptation, this could cause the additional parameters to be biased towards the major distribution due to the imbalanced distribution issue. We resolve this issue by stochastic adaptation, which allows an embedding space to consider both the generalizable features of a pre-trained model and adapted features, rather than relying solely on the adapted features. 
% This allows the embedding space to be unbiased to the major data distribution,
This alleviates bias in the embedding space toward the major data distribution,
while providing the capacity to learn knowledge specific to each dataset.
Our adapter has a bottleneck structure for parameter-efficiency and is connected in parallel with every transformer block. The adapter consists of a down-projection layer $W_\textrm{down} \in \mathbb{R}^{D \times r}$, a ReLU activation layer, and an up-projection layer $W_\textrm{up} \in \mathbb{R}^{r \times D}$, As shown in Fig.~\ref{fig:overview}, each transformer layer has two parallel adapters: one with the MSA block and one with the MLP block. Given input for the $\ell$-th transformer layer
and output of the $\ell$-th MSA layer, outputs of the adapters are produced as follows:
\begin{equation}
\begin{aligned}
\tilde{z}_{\ell}' &= \textrm{ReLU}(\textrm{LN}(z_{\ell - 1}) \cdot W_{\textrm{down}}') \cdot W_\textrm{up}' \\
\tilde{z}_{\ell}^{} &= \textrm{ReLU}(\textrm{LN}(z_{\ell}') \cdot W_{\textrm{down}}^{}) \cdot W_\textrm{up}^{},
\label{eq:transformer_Adapter}
\end{aligned}
\end{equation}
where $W_\textrm{down}'$ and $W_\textrm{up}'$ have the same shapes as $W_\textrm{down}$ and $W_\textrm{up}$, respectively.
The output features of the adapters are multiplied by random binary masks and combined with the outputs of the transformer blocks (\ie, MSA and MLP) through residual connections:
\begin{equation}
\begin{split}
z'_{\ell} &= \textrm{MSA}(\textrm{LN}(z_{\ell-1})) + z_{\ell-1} + \gamma'_{\ell} \cdot \tilde{z}_{\ell}', \\
z_{\ell} &= \textrm{MLP}(\textrm{LN}(z'_{\ell})) + z'_{\ell} + \gamma_{\ell} \cdot \tilde{z}_{\ell},
\label{eq:vit_with_adapter}
\end{split}
\end{equation}
where $\gamma_{\ell}'$ and $\gamma_{\ell}$ are independent variables drawn from $\textrm{Bernoulli}(p)$, and $p$ is the keep probability of the stochastic adapters.

\subsection{Conditional Prompt Learning}
\label{sec:prompt}

% \begin{wrapfigure}{t!}{0.33\textwidth}
% \vspace{-12mm}
% \includegraphics[width = 0.33\textwidth]{figs/attention_dataset2.pdf}
% \vspace{-4mm}
% \caption{
% The average similarity between input queries and
% prompts for each dataset. The $x$-axis represents the prompt index.
% } 
% \label{fig:attention}
% \vspace{2mm}
% \end{wrapfigure}

Conditional prompt learning aims to overcome the problem that the parameters shared by all data are inevitably dominated by the major data distribution.
We propose conditional prompt learning\footnote[1]{The prompt in ViT denotes the learnable input token parameters as in \cite{jia2022visual, wang2022learning,smith2022coda}} to learn more discriminative features for each dataset.
We assume that images within each dataset exhibit shared characteristics distinguished from those of other datasets. 
Our goal is to learn and leverage prompts relevant to the input data among the set of prompts through the attention mechanism.
% We aim to only learn and leverage prompts relevant to the input data among the set of prompts through the attention mechanism.
% Note that the prompt in ViT denotes the learnable input token parameters as in \cite{jia2022visual, wang2022learning,smith2022coda}.

To achieve this, a query feature that encodes the input image $x$ is first extracted. Query features should be able to grasp the data distribution of the input image and also require little computation to obtain it. Considering these requirements, we design a simple query feature for $x$ by using a pooling operation on its image patch embeddings $E$ in Sec.~\ref{sec:vit}:
\begin{equation}
q = \textrm{AvgPool}(E) + \textrm{MaxPool}(E), \quad q \in \mathbb{R}^{D}.
\end{equation}

\begin{figure} [t!]
\vspace{-1mm}
\centering
\includegraphics[width = 0.96\columnwidth]{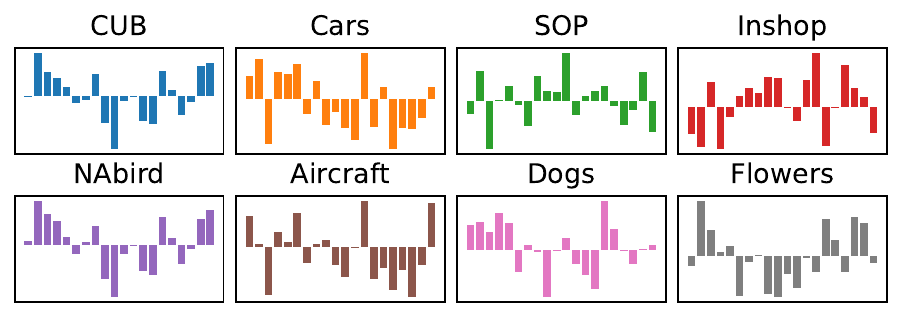}
\vspace{-3mm}
\caption{The average similarity between input queries and prompts for each dataset. The $x$-axis represents prompt indices.
} 
\vspace{-4mm}
\label{fig:attention}
\end{figure}

Then, we introduce a \emph{prompt pool}, a storage that contains prompts together with extra parameters for input-conditioning. $P_m \in \mathbb{R}^{N_p\times D}$ denote a prompt in the pool, where $N_p$ is the token length of a prompt, and then a prompt pool with $M$ prompts is given by:
\begin{equation}
\mathbf{P} = \{(P_1, K_1, A_1), \cdots ,(P_M, K_M, A_M)\},
\label{eq:prompt_pool}
\end{equation}
where $K_m \in \mathbb{R}^{D}$ denotes the key of a prompt, and $A_m \in \mathbb{R}^{D}$ is its feature attention vector, a learnable parameter emphasizing specific feature dimensions of the query vector. The query feature is element-wise multiplied with the feature attention vector to create an attended query, which is then paired with the prompt key for matching.
The weight vector is computed based on the cosine similarity between the attended query and the prompt key,
which is given by
\begin{equation}
\alpha_m = s(q \otimes A_m, K_m),
\label{eq:similarity}
\end{equation}
where $s(\cdot, \cdot)$ denotes the cosine similarity
between two vectors and $\otimes$ denotes element-wise product operation over the feature dimension.
The conditional prompt of input image $x$ is calculated as a weighted sum of prompts:
\begin{equation}
\hat{P} = \sum_{m=1}^M{\alpha_m P_m},
\label{eq:conditional_prompt}
\end{equation}
Finally, it is inserted into the input sequence of the transformer encoder:
\begin{equation}
z_0 = [e_\textrm{cls}, \hat{P}, E].
\label{eq:input_sequence_prompt}
\end{equation}

% Consequently, each prompt effectively conditions the input images based on their data distribution, which is supported by Fig.~\ref{fig:attention}. The relevant datasets such as CUB and NABird show a strong tendency to attend to similar prompts (correlation: 0.99). Conversely, unrelated datasets such as CUB and Inshop demonstrate a distinct preference for entirely different prompts (correlation: -0.67).

This process allows each prompt to condition images based on their specific data distributions, as depicted in Fig.~\ref{fig:attention}. Notably, while the CUB dataset demonstrates a strong tendency to align with the relevant NABird dataset, it prefers different prompts compared to the In-shop dataset.

% \begin{figure} [h!]
% \vspace{-2mm}
% \centering
% \includegraphics[width = 1.0\columnwidth]{figs/attention_dataset.pdf}
% \vspace{-7mm}
% \caption{The average similarity between input queries and prompts for each dataset. The $x$-axis represents prompt indices.
% } 
% \label{fig:attention}
% \vspace{-4mm}
% \end{figure}

% As a result, prompts are conditioned on individual input data rather than datasets. In addition, we found that samples of the same or closely related datasets tend to attend to similar prompts due to their shared features as supported by Fig.~\ref{fig:attention}. The relevant datasets such as CUB and NABird show a strong tendency to attend to similar prompts (correlation: 0.99), while unrelated datasets like CUB and Inshop select entirely different prompts (correlation: -0.67).

%%% EXPERIMENTS %%%%%%%%%%%%%%%%%%%%%%%%%%%%%%%%%%%%%%%%%
\begin{table*}[!t]
% \vspace{-1mm}
\setlength{\tabcolsep}{1pt}
\resizebox{\textwidth}{!}{
\fontsize{7.5}{9}\selectfont
\centering
\begin{tabularx}{1.0\textwidth}
    {
      p{0.145\textwidth}
      >{\centering\arraybackslash}p{0.11\textwidth}
      >{\centering\arraybackslash}p{0.065\textwidth}
      >{\centering\arraybackslash}p{0.065\textwidth}
      >{\centering\arraybackslash}p{0.065\textwidth}
      >{\centering\arraybackslash}p{0.065\textwidth}
      >{\centering\arraybackslash}p{0.065\textwidth}
      >{\centering\arraybackslash}p{0.065\textwidth}
      >{\centering\arraybackslash}p{0.065\textwidth}
      >{\centering\arraybackslash}p{0.065\textwidth}
      >{\centering\arraybackslash}X
      >{\centering\arraybackslash}X
      }
     \toprule
     
    {\multirow{1}{*}[-2.3mm]{Methods}}&
    {\multirow{1}{*}[-0.7mm]{Trainable}} &
    \multicolumn{8}{c}{\textbf{Dataset-Specific Accuracy}} & \multicolumn{2}{c}{\textbf{Universal Accuracy}} \\ [-0.3ex]   
    \cmidrule(lr){3-10} \cmidrule(lr){11-12}  & \multirow{1}{*}[0.7mm]{Params~(M)} & C{UB} & C{ars} & S{OP} & I{nShop} & N{ABird} & D{og} & F{lowers} & A{ircraft} & Unified &  Harmonic \\[-0.3ex] \midrule
    Zero-Shot  & 0.0 & 83.1 & 47.8 & 60.3 & 41.5 & 78.3 & \textbf{86.8} & \textbf{99.3} & 48.1 & 62.2 & 62.1 \\  [-0.3ex] \midrule
    \multicolumn{12}{c}{\textbf{(a) \textit{{Dataset-specific metric learning (Full fine-tuning)}}}} \\ [-0.3ex] \midrule
     Triplet~\cite{Schroff2015}  &{173.7} & 81.1 & 75.2 & 80.2 & 87.4 & 75.2 & 81.0 & 99.1 & 64.7 & 57.6 & 79.4\\
     Margin~\cite{sampling_matters}  &{173.7} & 79.4 & 78.0 & 79.8 & 86.0 & 74.6 & 80.3 & 99.0 & 66.8 & 58.1 & 79.6 \\
     MS~\cite{wang2019multi}  &{173.7} & 80.0 & 83.7 & 81.4 & 90.8 & 68.1 & 75.8 & 97.4 & 64.7 &{61.6} & 78.9 \\
     Proxy-Anchor~\cite{kim2020proxy} &{173.7} & 80.2 & 83.7 & \textbf{84.4} & 91.5 & 69.6 & 84.2 & 99.0 & 67.9 &{57.4} & {81.4} \\
     SoftTriple~\cite{Qian_2019_ICCV}  &{173.7} & 80.5 & 80.0 & 82.9 & 88.7 & 75.9 & 82.1 & \textbf{99.3} & 65.4 &{63.4} & {80.8} \\
     CosFace~\cite{wang2018cosface}  &{173.7} & 78.8 & 83.2 & 83.2 & 89.6 & 71.4 & 79.2 & 99.2 & 61.4 &{61.5} & {79.3} \\
     ArcFace~\cite{deng2019arcface}  &{173.7} & 76.8 & 79.4 & 83.4 & 90.3 & 61.0 & 76.1 & 99.2 & 60.0 &{58.8} & {76.2} \\
    CurricularFace~\cite{huang2020curricularface}  &{173.7} & 79.7 & 81.3 & 83.2 & 88.2 & 75.3 & 81.2 & 99.1 & 63.9 & 62.9 & 80.4\\
     Hyp~\cite{ermolov2022hyperbolic} &{173.7} &78.8 & 78.2 & 83.6 & 91.5 & 71.0 & 72.6 & 98.7 & 65.7 &{15.7} & {78.8} \\ \midrule
  \multicolumn{12}{c}{\textbf{(b) \textit{{Unified metric learning (Full fine-tuning)}}}} \\[-0.3ex] \midrule
     Triplet~\cite{Schroff2015}  &{21.7} &74.5 & 35.4 & 80.2 & 85.7 & 68.2 & 77.1 & 98.7 & 40.9 & 72.0 & 57.7\\
     Margin~\cite{sampling_matters}  &{21.7} & 72.5 & 36.7 & 80.0 & 84.1 & 67.4 & 74.8 & 98.5 & 40.4 & 71.6 & 57.4 \\
     MS~\cite{wang2019multi}  &{21.7} & 66.3 & 22.9 & 78.9 & 87.2 & 58.6 & 69.8 & 97.3 & 31.5 & 67.8 & 47.3 \\
     Proxy-Anchor~\cite{kim2020proxy}  &{21.7} & 77.2 & 73.1 & 83.7 & \textbf{91.9} & 71.5 & 78.1 & 96.4 & 62.7 & 77.9 & 71.0\\
     SoftTriple~\cite{Qian_2019_ICCV}  &{21.7} & 78.9 & 77.0 & 81.3 & 88.6 & 73.8 & 79.3 & 99.1 & 64.4 & 77.6 & 72.7 \\
     CosFace~\cite{wang2018cosface}  &{21.7} & 74.2 & 73.5 & 82.5 & 90.0 & 69.7 & 74.1 & 98.7 & 59.7 & 76.6 & 69.6\\
     ArcFace~\cite{deng2019arcface}  &{21.7} & 70.8 & 25.9 & 63.9 & 58.9 & 64.0 & 70.3 & 97.2 & 31.7 & 59.2 & 47.2 \\
     CurricularFace~\cite{huang2020curricularface}  &{21.7} & 78.3 & 77.9 & 82.0 & 89.1 & 73.0 & 79.3 & 99.1 & 65.6 & 77.9 & 79.5 \\
     Hyp~\cite{ermolov2022hyperbolic}  & {21.7} & 79.2 & 60.6 & 83.5 & 90.9 & 73.6 & 81.9 & 99.1 & 56.3 & 77.7 & 69.4\\   \midrule 
  \multicolumn{12}{c}{\textbf{(c) \textit{{Unified metric learning (Parameter-efficient fine-tuning)}}}} \\ [-0.3ex]\midrule
   % Linear Emb.  & {0.1~/~21.8} & 82.1 & 49.7 & 70.5 & 65.5 & 77.9 & \textbf{86.2} & 99.1 & 47.6 & 69.3 & 68.2 \\
   Linear Embedding  & {0.1} & 82.1 & 49.7 & 70.5 & 65.5 & 77.9 & {86.2} & 99.1 & 47.6 & 69.3 & 68.2 \\
     MLP-3 Embedding  & {5.3} & 57.5 & 29.7 &63.1& 63.2& 50.6& 64.5& 93.6& 32.8& 56.5& 50.3 \\
     VPT~\cite{jia2022visual} & {0.1} & 83.5 & 50.3 & 73.6 & 72.2 & 80.3 & {84.2} & 99.2 & 50.2 & 71.9 & 70.4 \\
     LoRA~\cite{hu2021lora}   & 2.4 & 77.0 & 70.9 & 81.3 & 86.2 & 70.8 & 79.1 & 98.9 & 59.7 & 76.1 & 76.5 \\
     AdaptFormer~\cite{chen2022adaptformer}  &{2.4} & 77.0 & 77.0 & 83.7 & 90.7 & 72.3 & 78.5 & 99.0 & 63.9 & 78.5 & 79.0 \\
   VPTSP-G~\cite{ren2024learning} & 124.8 & 75.7 & 41.0 & 77.9 & 78.3  & 70.4 & 81.6 & 98.8 & 41.7 & 71.4 & 64.6  \\ 
    \ccol Ours  & {\ccol \ccol 2.5} & \ccol \textbf{83.9} & \ccol \textbf{84.3}& \ccol {84.0} & {\ccol 89.8} & {\ccol \textbf{79.2}} & {\ccol 84.1} & {\ccol \textbf{99.3}} &  {\ccol \textbf{72.6}} &{\ccol \textbf{81.3}}& \ccol \textbf{84.1} \\
\bottomrule 
\end{tabularx}
}
\vspace{-1mm}
\caption{
Recall@1 of metric learning baselines and ours on the 8 datasets. Their network architecture is ViT-S/16~\cite{ViT} and uses 128 embedding dimensions except for the zero-shot model. Note that all methods are re-implemented and newly evaluated on the UML benchmark. VPTSP-G~\cite{ren2024learning} uses learnable prompts per class to generate semantic proxies, which require a substantial number of parameters due to the large number of classes in UML.
}
\label{tab:comparison_sota}
\vspace{-3mm}
\end{table*} 

\section{Experiments}

\subsection{Experimental Setup}
\label{sec:setup}

\noindent\textbf{Datasets.} In the UML setting, we employ a combination of eight datasets. These comprise four widely recognized benchmarks: CUB~\cite{CUB200}, Cars-196~\cite{krause20133d}, Standford Online Product (SOP)~\cite{songCVPR16}, and In-shop Clothes Retrieval (In-Shop)~\cite{DeepFashion}. Alongside these, we incorporate other four fine-grained datasets: NABirds~\cite{van2015building}, Dogs~\cite{khosla2011novel}, Flowers~\cite{nilsback2008automated}, and Aircraft~\cite{maji2013fine}. The overall dataset statistics are in Table~\ref{tab:dataset_stats}. The combined dataset encompasses 141,404 training images and 148,595 testing images. 
Notably, this dataset shows imbalanced data distributions, with a large portion of images from large-scale datasets such as SOP and In-Shop.

\begin{table*}[!t]

\centering
\resizebox{\textwidth}{!}{
\fontsize{7}{8.3}\selectfont
\begin{tabularx}{1.0\linewidth}
    {
      >{\centering\arraybackslash}X
      >{\centering\arraybackslash}X
      >{\centering\arraybackslash}X
      >{\centering\arraybackslash}X
      >{\centering\arraybackslash}p{0.041\textwidth}
      >{\centering\arraybackslash}p{0.041\textwidth}
      >{\centering\arraybackslash}p{0.041\textwidth}
      >{\centering\arraybackslash}p{0.041\textwidth}
      >{\centering\arraybackslash}p{0.041\textwidth}
      >{\centering\arraybackslash}p{0.041\textwidth}
      >{\centering\arraybackslash}p{0.041\textwidth}
      >{\centering\arraybackslash}p{0.041\textwidth}
      >{\centering\arraybackslash}p{0.041\textwidth}
      >{\centering\arraybackslash}p{0.06\textwidth}
      >{\centering\arraybackslash}p{0.06\textwidth}
      }
     \toprule
     \multicolumn{2}{c}{Prompt}& \multicolumn{2}{c}{Adapter} & {\multirow{1}{*}[-0.7mm]{Train}} & 
    \multicolumn{8}{c}{{Dataset-Specific Accuracy}} & \multicolumn{2}{c}{{Universal Accuracy}} \\ [-0.2ex]\cmidrule(lr){1-2}  \cmidrule(lr){3-4} \cmidrule(lr){6-13} \cmidrule(lr){14-15} \!\!\!\!Sing. & \!\!\!\!Cond. & Stat. & Stoc. & {\multirow{1}{*}[0.5mm]{Param.}} & C{UB} & C{ars} & \!\!S{OP} & \!\!\!\!\!I{nShop} & \!\!\!N{ABird} & D{og} & \!\!\!\!F{lowers} & \!\!\!A{ircraft} & Unified &  \!\!Harmonic \\[-0.2ex] \midrule
     % \multicolumn{12}{c}{{\textit{{Universal models} with ViT Backbone}}} \\ \midrule
    \redcheck & \xmark & \xmark & \xmark & 0.05M&\textbf{82.8} & 51.0 & 74.7 & 72.9 & 78.3 & {85.5} & 99.2 & 50.8 & 72.3 & 70.8 \\
    \xmark & \redcheck & \xmark & \xmark  & 0.13M & \textbf{82.8} &  \textbf{54.7} &  \textbf{76.8} &  \textbf{76.6} &  \textbf{78.7} &  \textbf{85.8} &  \textbf{99.3} &  \textbf{53.6} &  \textbf{74.0} &  \textbf{73.0} \\ \midrule
     \xmark & \xmark & \redcheck & \xmark & 2.41M & 74.5 & 81.3 & 83.7 & \textbf{90.4} & 74.5 & 81.3 & 99.0 & 66.3 & 79.4 & 80.8 \\
     \xmark & \xmark & \xmark & \redcheck &  2.41M &\textbf{83.6} &  \textbf{83.9} &  \textbf{83.8} &  89.9 &  \textbf{79.2} &  \textbf{84.6} &  \textbf{99.4} &  \textbf{71.9} &  \textbf{81.1} &  \textbf{83.9} \\ \midrule
    \redcheck & \xmark & \redcheck & \xmark & 2.41M & 79.6 & 80.0 & 83.8 & \textbf{90.3} & 73.7 & 81.0  & 99.0 & 65.4 & 79.3 & 80.5 \\ 
     \ccol \xmark & \ccol \redcheck &\ccol \xmark & \ccol\redcheck & \ccol  2.49M & \ccol \textbf{83.9} & \ccol\textbf{84.3}&  \ccol\textbf{84.0} & \ccol{ 89.8} &\ccol { \textbf{79.2}} &\ccol { \textbf{84.1}} & \ccol{ \textbf{99.3}} & \ccol { \textbf{72.6}} &{ \ccol\textbf{81.3}}& \ccol \textbf{84.1} \\ 
\bottomrule 
\end{tabularx}
}
\vspace{-1mm}
\caption{
Comparison between parameter-efficient transfer learning baseline with PUMA. ``Sing.'' denotes single prompt ($M=1$), ``Cond.'' denotes our conditional prompt ($M=20$), ``Stat.'' denotes adapter with $p=1$, and ``Stoc.'' denotes our conditional adapter with $p=0.5$.
}
\label{tab:ablation_component}
\vspace{-2mm}
\end{table*}

\noindent\textbf{Baselines.}
We benchmark ours against three distinct learning strategies. The models trained exclusively on individual datasets are termed \emph{dataset-specific metric learning}, while models trained on multiple datasets are termed \emph{unified metric learning} (UML).

\noindent\textbf{(a) Dataset-specific metric learning (Full fine-tuning)}: These models follow conventional metric learning protocols, where every parameter in the backbone and the embedding layer is updated. Each model is trained specifically for individual datasets, as depicted in Fig.~\ref{fig:comparison}(a).
For this approach, we employ a range of renowned metric learning losses, Triplet~\cite{Schroff2015}, Margin~\cite{sampling_matters}, MS~\cite{wang2019multi}, Proxy-Anchor~(PA)~\cite{kim2020proxy}, SoftTriple~\cite{Qian_2019_ICCV}, CosFace~\cite{wang2018cosface}, ArcFace~\cite{deng2019arcface}, CurricularFace~\cite{huang2020curricularface}, and Hyp~\cite{ermolov2022hyperbolic}. 

\noindent\textbf{(b) Unified metric learning (Full fine-tuning)}: The models are fully fine-tuned using the aforementioned losses, using a union of multiple datasets, as shown in Fig.~\ref{fig:comparison}(b).

\noindent\textbf{(c) Unified metric learning (Parameter-efficient fine-tuning)}: The models update a subset of backbone parameters or add new trainable
parameters to the backbone during the fine-tuning. While existing methods are designed for dataset-specific training, we implement and na\"ively apply them for UML for fair comparison with ours.
We explore two techniques focusing on the embedding layer: training solely the embedding layer with linear and 3-layered multilayer perceptron~(MLP-3 Embedding). We also consider three prominent parameter-efficient tuning strategies: VPT~\cite{jia2022visual}, LoRA~\cite{hu2021lora}, and AdaptFormer~\cite{chen2022adaptformer}. LoRA and AdaptFormer are scaled with the same parameters as ours. For training these models, we use the CurricularFace loss~\cite{huang2020curricularface} as the loss function.
Finally, we implement and compare VPT with Semantic Proxies using GRU fusion~(VPTSP-G)~\cite{ren2024learning} and PA loss, a state-of-the-art prompt tuning method by generating proxies in metric learning.

\noindent\textbf{Implementation Details.}
For fair comparisons, all models are evaluated using the same backbone, ViT-S/16~\cite{ViT} pre-trained on ImageNet-21K and then fine-tuned on ImageNet-1K~\cite{Imagenet}.
We change the size of its last layer to $128$, and $L_2$-normalize the output vector. 
We set the parameters for the stochastic adapter to $r=128$ and $p=0.5$, for the conditional prompt to $N_p=8$ and $M=20$. Unless otherwise specified, we adopt the CurricularFace~\cite{huang2020curricularface} as a loss function. We further ablate different loss functions and pre-trained backbones on our method and provide more implementation details in the appendix.

\noindent\textbf{Evaluation Protocol.}
In the main paper, we measure the performance using Recall@1, with additional results with R@$k$, MAP@R, and RP in the appendix.
We report the \emph{dataset-specific accuracy} using individual query and gallery for each dataset, and also calculate two kinds of \emph{universal accuracy}, the unified accuracy using unified query and gallery sets, and the harmonic mean of these individual accuracies. 
To evaluate the unified performance of dataset-specific models, we use an ensemble, averaging the embedding vectors of all dataset-specific models.

\subsection{Comparison with Existing Methods}
Table~\ref{tab:comparison_sota} shows Recall@1 performance with a total of eight datasets. We note that the total number of trainable parameters of dataset-specific models increases as the number of datasets increases.
\begin{itemize}[leftmargin=3mm]
\setlength\itemsep{0ex}
    \item Our results show that PUMA surpasses all compared data-specific models (Table~\ref{tab:comparison_sota}(a)) in terms of universal accuracy. Moreover, our method outperforms data-specific models in all cases except for In-Shop and Dog datasets, while not using hyperparameters selected for each respective dataset.
    Surprisingly, our method accomplishes this level of performance while \emph{using up to 69 times fewer trainable parameters} than previous techniques. This indicates that PUMA can be trained with limited resources and can easily be scaled up to a larger model and more datasets.
    Furthermore, even without emphasizing parameter efficiency, the outcomes highlight that our method can be a promising alternative to dataset-specific metric learning approaches.

    \item  Table~\ref{tab:comparison_sota}(b) shows the results of applying conventional full-finetuning methods to unified metric learning, revealing significant performance degradation on small datasets. Among various loss functions, CurricularFace loss~\cite{huang2020curricularface} performs better than others. PUMA consistently achieves high performance across all datasets, surpassing the best unified metric models in unified accuracy and harmonic mean accuracy by 3.4\% and 4.6\%, respectively.

    \item Among various PEFT methods (Table~\ref{tab:comparison_sota}(c)), only our method outperforms the majority of fully fine-tuned models. Models like linear embedding and VPT~\cite{jia2022visual}, which use fewer learnable parameters, notably underperform on datasets, such as Cars, SOP, and In-Shop. AdaptFormer~\cite{chen2022adaptformer} and LoRA~\cite{hu2021lora}, with similar parameters as our method, show biases towards large datasets akin to full fine-tuning. 
    The state-of-the-art method in metric learning, VPTSP-G~\cite{ren2024learning}, is inefficient in UML and performs poorly on smaller datasets. In contrast, PUMA shows its superior efficiency and effectiveness in handling diverse data distributions.
\end{itemize}

\subsection{Ablation Study}

\begin{table*}[!t]
\centering
% \vspace{-3mm}
\setlength{\tabcolsep}{1pt}
\resizebox{\textwidth}{!}{
\fontsize{7}{8.1}\selectfont
\begin{tabularx}{1.0\textwidth}
    {
      >{\arraybackslash}p{0.17\textwidth}
      >{\centering\arraybackslash}p{0.088\textwidth}
      >{\centering\arraybackslash}p{0.088\textwidth}
      >{\centering\arraybackslash}p{0.088\textwidth}
      >{\centering\arraybackslash}p{0.088\textwidth}
      >{\centering\arraybackslash}p{0.088\textwidth}
      >{\centering\arraybackslash}p{0.088\textwidth}
      >{\centering\arraybackslash}p{0.088\textwidth}
      >{\centering\arraybackslash}p{0.088\textwidth}
      >{\centering\arraybackslash}p{0.088\textwidth}
      }
     \toprule
     {Methods}& C{UB} & C{ars} & \!S{OP} & \!\!\!I{nShop} & \!\!\!N{ABird} & D{og} & \!\!\!F{lowers} & \!\!A{ircraft} & \!\!Harmonic \\ \midrule
     % \multicolumn{10}{l}{{(a) \textit{\textbf{Models by Dataset-specific Training}}}} \\ \midrule
    PA (Spec.) &80.2&83.7&84.4&91.5&69.6&84.2& 99.0& 67.9& 81.4\\
    CurricularFace (Spec.) &79.7&81.3&83.2&88.2& 75.3&81.2&99.1&63.9& 80.4\\
     Ours (Spec.) & 81.7& 83.9& \textbf{85.2}& \textbf{89.9}& 77.0& 82.5& \textbf{99.5}& 68.9&82.7 \\ \midrule
     % \multicolumn{10}{l}{{(b) \textit{\textbf{Models by Universal Training}}}} \\ \midrule
PA (Univ.) & 77.2~\newred{(-3.0)} & 73.1~\newred{(-10.6)}& 83.7~\textbf{\newred{(-0.7)}}& 91.9~\sytodo{(+0.4)}& 71.5~\sytodo{(+1.9)}& 78.1~\newred{(-6.1)}& 96.4~\newred{(-2.6)}& 62.7~\newred{(-5.2)}& 71.0~\newred{(-10.4)}\\
CurricularFace (Univ.) & 78.3~\newred{(-1.4)} &77.9~\newred{(-3.4)} &82.0~\newred{(-1.2)} &89.1~\textbf{\sytodo{(+0.9)}}& 73.0~\newred{(-2.3)} &79.3~\newred{(-1.9)} &99.1~{(+0.0)} &65.6~\newred{(-1.7)} &79.5~\newred{(-0.9)}\\
\ccol Ours (Univ.) & \ccol {\textbf{83.9}~\sytodo{\textbf{(+2.2)}}} &\ccol \textbf{84.3}~\sytodo{\textbf{{(+0.4)}}} &\ccol {84.0~\newred{(-1.2)}} &\ccol {89.8}~\newred{(-0.1)} &\ccol \textbf{79.2}~\sytodo{\textbf{{(+2.2)}}} &\ccol {\textbf{84.1}}~\textbf{\sytodo{(1.6)}} &\ccol 99.3~\newred{\textbf{{(-0.2)}}} &\ccol \textbf{72.6}~\sytodo{\textbf{{(+3.7)}}} &\ccol \textbf{84.1}~\sytodo{\textbf{{(+1.4)}}}    \\  
% 85.0. 84.0, 84.5, 909. 80.8, 82.8, 99.4, 72.7, 82.1, 84.4
\bottomrule 
\end{tabularx}
}
\vspace{-1mm}
\caption{Comparison of performance between dataset-specific training (\ie., {Spec.}) and unified training (\ie, {Univ.}). \newred{(-)} and \sytodo{(+)} denote degradation and enhancement in performance achieved when transitioning from dataset-specific training to unified training, respectively.
}
\label{subtab:specific_vs_universal}
\vspace{-1mm}
\end{table*} 
\begin{table*}[!t]
\centering
\fontsize{7}{8.1}\selectfont
\resizebox{\textwidth}{!}{
\setlength{\tabcolsep}{2pt}
\begin{tabularx}{1.0\linewidth}
    {
      >{\arraybackslash}p{0.11\textwidth}
      >{\centering\arraybackslash}p{0.12\textwidth}
      >{\centering\arraybackslash}p{0.21\textwidth}
      >{\centering\arraybackslash}p{0.04\textwidth}
      >{\centering\arraybackslash}p{0.04\textwidth}
      >{\centering\arraybackslash}p{0.04\textwidth}
      >{\centering\arraybackslash}p{0.048\textwidth}
      >{\centering\arraybackslash}p{0.048\textwidth}
      >{\centering\arraybackslash}p{0.04\textwidth}
      >{\centering\arraybackslash}p{0.048\textwidth}
      >{\centering\arraybackslash}p{0.048\textwidth}
      >{\centering\arraybackslash}X
      >{\centering\arraybackslash}X
      }
     \toprule
     {Methods}&
    {Arch.} & Pre-training & C{UB} & C{ars} & S{OP} & I{nShop} & N{ABird} & D{og} & F{lowers} & A{ircraft} & U{nif.} & H{arm.} \\ \midrule

     Zero-Shot  & { {CLIP-B/32}$^{512}$} &  LAION-2B (Language Sup.) & 78.3 & 41.3 & 55.0 & 34.9 & 73.2 & {85.0} & \textbf{99.1} & 42.9 & 57.3 & 56.1\\
     CurricularFace &{{CLIP-B/32}$^{128}$}  & LAION-2B (Language Sup.) &74.1& 78.7 & 80.8 & 88.4 & 67.1 & 80.8  &  98.5& 64.5&76.3 & 77.8  \\
    \ccol Ours& {\ccol {CLIP-B/32}$^{128}$} &\ccol  LAION-2B (Language Sup.) & \ccol \textbf{81.1} & \ccol \textbf{85.6} & \ccol \textbf{83.9} & {\ccol \textbf{90.5}} & \ccol \textbf{75.2} & \ccol \textbf{87.0} & \ccol {99.0} &  {\ccol \textbf{72.0}} &\ccol \textbf{81.0}& \ccol \textbf{83.6} \\ \midrule
    
     Zero-Shot  &{{ViT-S/16}$^{384}$} & IN-21K \& IN-1K  (Sup.) & 83.1 & 47.8 & 60.3 & 41.5 & 78.3 & \textbf{86.8} & \textbf{99.3} & 48.1 & 62.2 & 62.1 \\
     CurricularFace  &{{ViT-S/16}$^{128}$} & IN-21K \& IN-1K  (Sup.) & 78.3 & 77.9 & 82.0 & 89.1 & 73.0 & 79.3 & 99.1 & 65.6 & 77.9 & 79.5 \\
    \ccol Ours & \ccol {ViT-S/16$^{128}$} & \ccol IN-21K \& IN-1K  (Sup.) & \ccol \textbf{83.9} &\ccol \textbf{84.3} &\ccol \textbf{84.0} &\ccol \textbf{89.8} &\ccol \textbf{79.2} &\ccol {84.1} &\ccol \textbf{99.3} &\ccol \textbf{72.6}&\ccol  \textbf{81.3} &\ccol \textbf{84.1} \\ \midrule
     % PA &\multicolumn{1}{c}{{ViT-S/16}$^{512}$} &21.9M & 78.0 & 76.5 & 85.0 & 92.2 & 72.8 & 79.2 & 96.7 & 67.0 & 79.4 & 73.2\\
     % CosFace &{{ViT-S}$^{512}$} &21.9M & 74.7 & 77.7 & 83.8 & 90.5 & 71.0 & 75.4 & 99.0 & 63.4 & 78.1 & 71.6\\     
     Zero-Shot  &{{DeiT-S/16}$^{384}$} & IN-1K (Sup.) & 57.1 & 40.6 & 52.8 & 31.5 & 46.8 & \textbf{87.9} & 88.6 & 34.0 & 50.2 & 48.1\\
     CurricularFace &{{DeiT-S/16}$^{128}$}  & IN-1K (Sup.) & 69.7 & 76.9 & 79.4 & 88.4 & 62.3 & 81.7 & 94.9 & 63.2 & 74.3 & 75.5 \\
    \ccol Ours& {\ccol {DeiT-S/16}$^{128}$} &\ccol  IN-1K (Sup.) & \ccol \textbf{75.8} & \ccol \textbf{83.1} & \ccol \textbf{82.0} & {\ccol \textbf{90.4}} & {\ccol \textbf{69.9}} & {\ccol {86.8}} & {\ccol \textbf{96.4}} &  {\ccol \textbf{70.1}} &{\ccol \textbf{78.4}}& {\ccol \textbf{80.8}} \\ \midrule
     Zero-Shot  & {{DINO-S/16}$^{384}$} & IN-1K (Self-Sup.) & 63.8 & 38.9 & 57.6 & 39.1 & 55.2 & {79.6} & 94.7 & 52.8 & 54.9 & 55.3\\
     CurricularFace &{{DINO-S/16}$^{128}$}  & IN-1K (Self-Sup.) &61.3& 68.7 & 79.0 & 88.2 & 53.6 & 61.4  &  92.2& 56.6&70.1 & 67.6  \\
    \ccol Ours& {\ccol {DINO-S/16}$^{128}$} &\ccol  IN-1K (Self-Sup.) & \ccol \textbf{78.1} & \ccol \textbf{84.0} & \ccol \textbf{82.3} & {\ccol \textbf{89.6}} & {\ccol \textbf{72.1}} & {\ccol \textbf{82.0}} & {\ccol \textbf{96.8}} &  {\ccol \textbf{73.3}} &{\ccol \textbf{78.8}}& {\ccol \textbf{81.6}} 
    \\ 
\bottomrule 
\end{tabularx}
}
\vspace{-1mm}
\caption{Recall@1 of our method compared to the best UML baseline model (\ie, CurricularFace) using backbones with different pre-training strategies. Superscripts denote embedding dimensions.
}
\label{tab:diff_arch}
\vspace{-3mm}
\end{table*}

\noindent\textbf{Impact of unique design of components.} 
Table~\ref{tab:ablation_component} shows the effectiveness of conditional prompts and stochastic adapters, compared to conventional prompt tuning and adapters. Compared to the single prompt, the introduction of conditional prompts significantly improves performance, especially in datasets where color is not a discriminative feature, such as Cars and Aircraft. It suggests that conditional prompts facilitate the learning of distinctive dataset characteristics. Stochastic adapters, in contrast to static ones, provide a more balanced improvement across all datasets, avoiding bias towards larger datasets such as SOP and InShop. Notably, combining existing methods yields minimal gains or often degradation, but the combination of the proposed modules boosts overall performance. 

\noindent\textbf{Dataset-specific vs. Unified Training.}
Table~\ref{subtab:specific_vs_universal} presents a comparison between dataset-specific training and unified training across methods. A distinct pattern emerges: while most methods exhibit a decrease in performance with unified training, our method notably shows significant improvements. This difference suggests that previous methods tend to be biased towards larger datasets with unified training, leading to a drop in performance. In contrast, our approach successfully captures and leverages shared characteristics across diverse datasets, resulting in a remarkable enhancement in performance. Such an increase not only enables an evaluation across different data distributions but also suggests that our method could be a pivotal approach to universally boost performance across various datasets.

\noindent\textbf{Different Pre-trained Backbones.}
We analyze the effect of backbone architectures and pre-training strategies on retrieval performance, as summarized in Table~\ref{tab:diff_arch}, comparing our method to the best baseline, CurricularFace. The analysis highlights three key observations:
(i) \emph{Pre-training strategies strongly influence retrieval performance but are not sufficient on their own.} Models trained on larger datasets with language or supervised pre-training (e.g., LAION-2B or IN-21K) generally perform better than those trained on smaller datasets (e.g., IN-1K), but generalization to distinct downstream tasks remains challenging without fine-tuning. Even large-scale pre-trained models like CLIP struggle with unified metric learning, highlighting persistent challenges.
(ii) \emph{Our method consistently outperforms the baseline across all pre-training strategies.} The performance gap increases as pre-training quality decreases, showing the ability of our method to enhance retrieval robustness, even with weaker pre-training (e.g., IN-1K self-supervised).
(iii) \emph{Baselines show growing bias towards larger datasets with weaker pre-training}, likely due to reduced representation generalizability. In contrast, our method effectively retains and adapts pre-trained knowledge, achieving uniform improvements across datasets.

%%% CONCLUSION %%%%%%%%%%%%%%%%%%%%%%%%%%%%%%%%%%%%%%%%%%
\section{Conclusion}
% Previous work on deep metric learning has focused primarily on developing dataset-specific models.
% However, this approach is limited in terms of scalability since real-world applications usually accommodate diverse data distributions. 
Previous deep metric learning studies have focused on dataset-specific models, which limit scalability for real-world applications with diverse data distributions.
In this paper, we have investigated UML, which enables a single model to manage multiple heterogeneous data distributions. 
In the UML setting, existing metric learning baselines suffer from imbalanced data distributions. 
To address this issue, we have proposed parameter-efficient tuning that is simple, lightweight, and achieves state-of-the-art performance using only a single model. 
We believe our work will facilitate future investigations into bridging the gap between metric learning and real-world applications. 

% \vspace{0.1cm}
% {\small \noindent \textbf{Acknowledgement:} 
% This work was supported by NRF grants (NRF-2021R1A2C3012728--30\%, NRF-2018R1A5A1060031--30\%, RS-2024-00341514--25\%) and IITP grants (RS-2019-II191906--10\%, Artificial Intelligence Graduate School Program - POSTECH, RS-2019-II190079--5\%, Artificial Intelligence Graduate School Program - Korea University) funded by Ministry of Science and ICT, Korea.
% }

% \vspace{0.1cm}
% \noindent\textbf{Acknowledgment.} This work was supported by Samsung Electronics Co., Ltd (XXXXXXXXXX), the NRF grants (NRF-2021R1A2C3012728--30\%, RS-2024-00341514--25\%), and the IITP grants (RS-2019-II191906--20\%, RS-2024-00457882--20\%, AI Graduate School Program at POSTECH, 2019-II190079--5\%, AI Graduate School Program at Korea University) funded by the Korea government (MSIT).

\noindent\textbf{Acknowledgment.} This work was supported by Samsung Advanced Institute of Technology, Samsung Electronics Co., Ltd., the NRF grants (NRF-2021R1A2C3012728--30\%, RS-2024-00341514--25\%), and the IITP grants (RS-2019-II191906--20\%, RS-2024-00457882--20\%, AI Graduate School Program at POSTECH, 2019-II190079--5\%, AI Graduate School Program at Korea University) funded by the Korea government (MSIT).

\bibliographystyle{ieee_fullname}
\bibliography{cvlab_kwak}

\begin{thebibliography}{10}\itemsep=-1pt

\bibitem{almazan2022granularity}
Jon Almaz{\'a}n, Byungsoo Ko, Geonmo Gu, Diane Larlus, and Yannis Kalantidis.
\newblock Granularity-aware adaptation for image retrieval over multiple tasks.
\newblock In {\em European Conference on Computer Vision}, pages 389--406. Springer, 2022.

\bibitem{bell2020groknet}
Sean Bell, Yiqun Liu, Sami Alsheikh, Yina Tang, Edward Pizzi, M Henning, Karun Singh, Omkar Parkhi, and Fedor Borisyuk.
\newblock Groknet: Unified computer vision model trunk and embeddings for commerce.
\newblock In {\em Proceedings of the 26th ACM SIGKDD international conference on knowledge discovery \& data mining}, 2020.

\bibitem{Bromley1994}
Jane Bromley, Isabelle Guyon, Yann Lecun, Eduard Säckinger, and Roopak Shah.
\newblock Signature verification using a "siamese" time delay neural network.
\newblock In {\em Proc. Neural Information Processing Systems (NeurIPS)}, 1994.

\bibitem{chen2022adaptformer}
Shoufa Chen, Chongjian Ge, Zhan Tong, Jiangliu Wang, Yibing Song, Jue Wang, and Ping Luo.
\newblock Adaptformer: Adapting vision transformers for scalable visual recognition.
\newblock In {\em Proc. Neural Information Processing Systems (NeurIPS)}, 2022.

\bibitem{chen2020simple}
Ting Chen, Simon Kornblith, Mohammad Norouzi, and Geoffrey Hinton.
\newblock A simple framework for contrastive learning of visual representations.
\newblock In {\em Proc. International Conference on Machine Learning (ICML)}, 2020.

\bibitem{Chopra2005}
S. Chopra, R. Hadsell, and Y. LeCun.
\newblock Learning a similarity metric discriminatively, with application to face verification.
\newblock In {\em Proc. IEEE Conference on Computer Vision and Pattern Recognition (CVPR)}, 2005.

\bibitem{Imagenet}
Jia Deng, Wei Dong, Richard Socher, Li-Jia Li, Kai Li, and Li Fei-Fei.
\newblock {ImageNet:} a large-scale hierarchical image database.
\newblock In {\em Proc. IEEE Conference on Computer Vision and Pattern Recognition (CVPR)}, 2009.

\bibitem{deng2019arcface}
Jiankang Deng, Jia Guo, Niannan Xue, and Stefanos Zafeiriou.
\newblock Arcface: Additive angular margin loss for deep face recognition.
\newblock In {\em Proc. IEEE Conference on Computer Vision and Pattern Recognition (CVPR)}, 2019.

\bibitem{ViT}
Alexey Dosovitskiy, Lucas Beyer, Alexander Kolesnikov, Dirk Weissenborn, Xiaohua Zhai, Thomas Unterthiner, Mostafa Dehghani, Matthias Minderer, Georg Heigold, Sylvain Gelly, et~al.
\newblock An image is worth 16x16 words: Transformers for image recognition at scale.
\newblock In {\em Proc. International Conference on Learning Representations (ICLR)}, 2021.

\bibitem{ermolov2022hyperbolic}
Aleksandr Ermolov, Leyla Mirvakhabova, Valentin Khrulkov, Nicu Sebe, and Ivan Oseledets.
\newblock Hyperbolic vision transformers: Combining improvements in metric learning.
\newblock In {\em Proc. IEEE Conference on Computer Vision and Pattern Recognition (CVPR)}, 2022.

\bibitem{Hadsell2006}
R. Hadsell, S. Chopra, and Y. LeCun.
\newblock Dimensionality reduction by learning an invariant mapping.
\newblock In {\em Proc. IEEE Conference on Computer Vision and Pattern Recognition (CVPR)}, 2006.

\bibitem{he2021towards}
Junxian He, Chunting Zhou, Xuezhe Ma, Taylor Berg-Kirkpatrick, and Graham Neubig.
\newblock Towards a unified view of parameter-efficient transfer learning.
\newblock {\em arXiv preprint arXiv:2110.04366}, 2021.

\bibitem{he2020momentum}
Kaiming He, Haoqi Fan, Yuxin Wu, Saining Xie, and Ross Girshick.
\newblock Momentum contrast for unsupervised visual representation learning.
\newblock In {\em Proc. IEEE Conference on Computer Vision and Pattern Recognition (CVPR)}, 2020.

\bibitem{hou2022towards}
Yupeng Hou, Shanlei Mu, Wayne~Xin Zhao, Yaliang Li, Bolin Ding, and Ji-Rong Wen.
\newblock Towards universal sequence representation learning for recommender systems.
\newblock In {\em Proceedings of the 28th ACM SIGKDD Conference on Knowledge Discovery and Data Mining}, 2022.

\bibitem{houlsby2019parameter}
Neil Houlsby, Andrei Giurgiu, Stanislaw Jastrzebski, Bruna Morrone, Quentin De~Laroussilhe, Andrea Gesmundo, Mona Attariyan, and Sylvain Gelly.
\newblock Parameter-efficient transfer learning for nlp.
\newblock In {\em Proc. International Conference on Machine Learning (ICML)}. PMLR, 2019.

\bibitem{hu2021lora}
Edward~J Hu, Yelong Shen, Phillip Wallis, Zeyuan Allen-Zhu, Yuanzhi Li, Shean Wang, Lu Wang, and Weizhu Chen.
\newblock Lora: Low-rank adaptation of large language models.
\newblock {\em arXiv preprint arXiv:2106.09685}, 2021.

\bibitem{huang2020embedding}
Jui-Ting Huang, Ashish Sharma, Shuying Sun, Li Xia, David Zhang, Philip Pronin, Janani Padmanabhan, Giuseppe Ottaviano, and Linjun Yang.
\newblock Embedding-based retrieval in facebook search.
\newblock In {\em Proceedings of the 26th ACM SIGKDD International Conference on Knowledge Discovery \& Data Mining}, 2020.

\bibitem{huang2020curricularface}
Yuge Huang, Yuhan Wang, Ying Tai, Xiaoming Liu, Pengcheng Shen, Shaoxin Li, Jilin Li, and Feiyue Huang.
\newblock Curricularface: adaptive curriculum learning loss for deep face recognition.
\newblock In {\em Proc. IEEE Conference on Computer Vision and Pattern Recognition (CVPR)}, 2020.

\bibitem{jia2022visual}
Menglin Jia, Luming Tang, Bor-Chun Chen, Claire Cardie, Serge Belongie, Bharath Hariharan, and Ser-Nam Lim.
\newblock Visual prompt tuning.
\newblock In {\em Proc. European Conference on Computer Vision (ECCV)}. Springer, 2022.

\bibitem{khosla2011novel}
Aditya Khosla, Nityananda Jayadevaprakash, Bangpeng Yao, and Fei-Fei Li.
\newblock Novel dataset for fine-grained image categorization: Stanford dogs.
\newblock In {\em Proc. CVPR workshop on fine-grained visual categorization (FGVC)}. Citeseer, 2011.

\bibitem{kim2020proxy}
Sungyeon Kim, Dongwon Kim, Minsu Cho, and Suha Kwak.
\newblock Proxy anchor loss for deep metric learning.
\newblock In {\em Proc. IEEE Conference on Computer Vision and Pattern Recognition (CVPR)}, 2020.

\bibitem{kim2019deep}
Sungyeon Kim, Minkyo Seo, Ivan Laptev, Minsu Cho, and Suha Kwak.
\newblock Deep metric learning beyond binary supervision.
\newblock In {\em Proc. IEEE Conference on Computer Vision and Pattern Recognition (CVPR)}, 2019.

\bibitem{krause20133d}
Jonathan Krause, Michael Stark, Jia Deng, and Li Fei-Fei.
\newblock 3d object representations for fine-grained categorization.
\newblock In {\em Proceedings of the IEEE International Conference on Computer Vision Workshops}, pages 554--561, 2013.

\bibitem{lester2021power}
Brian Lester, Rami Al-Rfou, and Noah Constant.
\newblock The power of scale for parameter-efficient prompt tuning.
\newblock In {\em Proceedings of the 2021 Conference on Empirical Methods in Natural Language Processing}, 2021.

\bibitem{li2021prefix}
Xiang~Lisa Li and Percy Liang.
\newblock Prefix-tuning: Optimizing continuous prompts for generation.
\newblock {\em arXiv preprint arXiv:2101.00190}, 2021.

\bibitem{DeepFashion}
Ziwei Liu, Ping Luo, Shi Qiu, Xiaogang Wang, and Xiaoou Tang.
\newblock Deepfashion: Powering robust clothes recognition and retrieval with rich annotations.
\newblock In {\em Proc. IEEE Conference on Computer Vision and Pattern Recognition (CVPR)}, 2016.

\bibitem{ma2020probabilistic}
Chen Ma, Liheng Ma, Yingxue Zhang, Ruiming Tang, Xue Liu, and Mark Coates.
\newblock Probabilistic metric learning with adaptive margin for top-k recommendation.
\newblock In {\em Proceedings of the 26th ACM SIGKDD International Conference on knowledge discovery \& data mining}, 2020.

\bibitem{maji2013fine}
Subhransu Maji, Esa Rahtu, Juho Kannala, Matthew Blaschko, and Andrea Vedaldi.
\newblock Fine-grained visual classification of aircraft.
\newblock {\em arXiv preprint arXiv:1306.5151}, 2013.

\bibitem{movshovitz2017no}
Yair Movshovitz-Attias, Alexander Toshev, Thomas~K Leung, Sergey Ioffe, and Saurabh Singh.
\newblock No fuss distance metric learning using proxies.
\newblock In {\em Proc. IEEE International Conference on Computer Vision (ICCV)}, 2017.

\bibitem{nilsback2008automated}
Maria-Elena Nilsback and Andrew Zisserman.
\newblock Automated flower classification over a large number of classes.
\newblock In {\em 2008 Sixth Indian Conference on Computer Vision, Graphics \& Image Processing}, pages 722--729. IEEE, 2008.

\bibitem{pfeiffer2020adapterfusion}
Jonas Pfeiffer, Aishwarya Kamath, Andreas R{\"u}ckl{\'e}, Kyunghyun Cho, and Iryna Gurevych.
\newblock Adapterfusion: Non-destructive task composition for transfer learning.
\newblock {\em arXiv preprint arXiv:2005.00247}, 2020.

\bibitem{Qian_2019_ICCV}
Qi Qian, Lei Shang, Baigui Sun, Juhua Hu, Hao Li, and Rong Jin.
\newblock Softtriple loss: Deep metric learning without triplet sampling.
\newblock In {\em Proc. IEEE International Conference on Computer Vision (ICCV)}, 2019.

\bibitem{multi_domain}
S.-A. Rebuffi, H. Bilen, and A. Vedaldi.
\newblock Learning multiple visual domains with residual adapters.
\newblock In {\em Proc. Neural Information Processing Systems (NeurIPS)}, 2017.

\bibitem{ren2024learning}
Li Ren, Chen Chen, Liqiang Wang, and Kien Hua.
\newblock Learning semantic proxies from visual prompts for parameter-efficient fine-tuning in deep metric learning.
\newblock In {\em Proc. International Conference on Learning Representations (ICLR)}, 2024.

\bibitem{Schroff2015}
Florian Schroff, Dmitry Kalenichenko, and James Philbin.
\newblock {FaceNet: A unified embedding for face recognition and clustering}.
\newblock In {\em Proc. IEEE Conference on Computer Vision and Pattern Recognition (CVPR)}, 2015.

\bibitem{smith2022coda}
James~Seale Smith, Leonid Karlinsky, Vyshnavi Gutta, Paola Cascante-Bonilla, Donghyun Kim, Assaf Arbelle, Rameswar Panda, Rogerio Feris, and Zsolt Kira.
\newblock Coda-prompt: Continual decomposed attention-based prompting for rehearsal-free continual learning.
\newblock {\em arXiv preprint arXiv:2211.13218}, 2022.

\bibitem{Sohn_nips2016}
Kihyuk Sohn.
\newblock Improved deep metric learning with multi-class n-pair loss objective.
\newblock In {\em Proc. Neural Information Processing Systems (NeurIPS)}, 2016.

\bibitem{songCVPR17}
Hyun~Oh Song, Stefanie Jegelka, Vivek Rathod, and Kevin Murphy.
\newblock Deep metric learning via facility location.
\newblock In {\em Proc. IEEE Conference on Computer Vision and Pattern Recognition (CVPR)}, 2017.

\bibitem{songCVPR16}
Hyun~Oh Song, Yu Xiang, Stefanie Jegelka, and Silvio Savarese.
\newblock Deep metric learning via lifted structured feature embedding.
\newblock In {\em Proc. IEEE Conference on Computer Vision and Pattern Recognition (CVPR)}, 2016.

\bibitem{tang2019msuru}
Yina Tang, Fedor Borisyuk, Siddarth Malreddy, Yixuan Li, Yiqun Liu, and Sergey Kirshner.
\newblock Msuru: Large scale e-commerce image classification with weakly supervised search data.
\newblock In {\em Proceedings of the 25th ACM SIGKDD International Conference on Knowledge Discovery \& Data Mining}, 2019.

\bibitem{teh2020proxynca++}
Eu~Wern Teh, Terrance DeVries, and Graham~W Taylor.
\newblock Proxynca++: Revisiting and revitalizing proxy neighborhood component analysis.
\newblock In {\em European Conference on Computer Vision (ECCV)}. Springer, 2020.

\bibitem{van2015building}
Grant Van~Horn, Steve Branson, Ryan Farrell, Scott Haber, Jessie Barry, Panos Ipeirotis, Pietro Perona, and Serge Belongie.
\newblock Building a bird recognition app and large scale dataset with citizen scientists: The fine print in fine-grained dataset collection.
\newblock In {\em Proc. IEEE Conference on Computer Vision and Pattern Recognition (CVPR)}, 2015.

\bibitem{wang2018cosface}
Hao Wang, Yitong Wang, Zheng Zhou, Xing Ji, Dihong Gong, Jingchao Zhou, Zhifeng Li, and Wei Liu.
\newblock Cosface: Large margin cosine loss for deep face recognition.
\newblock In {\em Proc. IEEE Conference on Computer Vision and Pattern Recognition (CVPR)}, 2018.

\bibitem{Wang2014}
Jiang Wang, Yang Song, T. Leung, C. Rosenberg, Jingbin Wang, J. Philbin, Bo Chen, and Ying Wu.
\newblock Learning fine-grained image similarity with deep ranking.
\newblock In {\em Proc. IEEE Conference on Computer Vision and Pattern Recognition (CVPR)}, 2014.

\bibitem{wang2019multi}
Xun Wang, Xintong Han, Weilin Huang, Dengke Dong, and Matthew~R Scott.
\newblock Multi-similarity loss with general pair weighting for deep metric learning.
\newblock In {\em Proc. IEEE Conference on Computer Vision and Pattern Recognition (CVPR)}, 2019.

\bibitem{wang2019ranked}
Xinshao Wang, Yang Hua, Elyor Kodirov, Guosheng Hu, Romain Garnier, and Neil~M Robertson.
\newblock Ranked list loss for deep metric learning.
\newblock In {\em Proc. IEEE Conference on Computer Vision and Pattern Recognition (CVPR)}, 2019.

\bibitem{wang2022learning}
Zifeng Wang, Zizhao Zhang, Chen-Yu Lee, Han Zhang, Ruoxi Sun, Xiaoqi Ren, Guolong Su, Vincent Perot, Jennifer Dy, and Tomas Pfister.
\newblock Learning to prompt for continual learning.
\newblock In {\em Proc. IEEE Conference on Computer Vision and Pattern Recognition (CVPR)}, 2022.

\bibitem{CUB200}
P. Welinder, S. Branson, T. Mita, C. Wah, F. Schroff, S. Belongie, and P. Perona.
\newblock {Caltech-UCSD Birds 200}.
\newblock Technical Report CNS-TR-2010-001, California Institute of Technology, 2010.

\bibitem{sampling_matters}
Chao-Yuan Wu, R. Manmatha, Alexander~J. Smola, and Philipp Krahenbuhl.
\newblock Sampling matters in deep embedding learning.
\newblock In {\em Proc. IEEE International Conference on Computer Vision (ICCV)}, 2017.

\bibitem{uned}
Nikolaos-Antonios Ypsilantis, Kaifeng Chen, Bingyi Cao, M{\'a}rio Lipovsk{\`y}, Pelin Dogan-Sch{\"o}nberger, Grzegorz Makosa, Boris Bluntschli, Mojtaba Seyedhosseini, Ond{\v{r}}ej Chum, and Andr{\'e} Araujo.
\newblock Towards universal image embeddings: A large-scale dataset and challenge for generic image representations.
\newblock In {\em Proc. IEEE International Conference on Computer Vision (ICCV)}, 2023.

\bibitem{zhai2018classification}
Andrew Zhai and Hao-Yu Wu.
\newblock Classification is a strong baseline for deep metric learning.
\newblock {\em arXiv preprint arXiv:1811.12649}, 2018.

\bibitem{zhang2018visual}
Yanhao Zhang, Pan Pan, Yun Zheng, Kang Zhao, Yingya Zhang, Xiaofeng Ren, and Rong Jin.
\newblock Visual search at alibaba.
\newblock In {\em Proceedings of the 24th ACM SIGKDD international conference on knowledge discovery \& data mining}, 2018.

\end{thebibliography}

% \clearpage
% \appendix
% \input{Sections/_7_supple}

\end{document}